\documentclass{article}
\usepackage[final,sglblindworkshop,nonatbib]{neurips_2026}
\workshoptitle{AI4Mat 2026}

\usepackage[utf8]{inputenc}
\usepackage[T1]{fontenc}
\usepackage{amsmath,amssymb}
\usepackage{microtype}
\usepackage{booktabs}
\usepackage{array}
\usepackage{xcolor}
\usepackage{hyperref}
\hypersetup{hidelinks}
\usepackage{graphicx}
\usepackage{tikz}
\usetikzlibrary{positioning,shapes.geometric,arrows.meta,fit,calc}
\usepackage{enumitem}
\usepackage{placeins}

\title{Diagnostic Accuracy Is Not Deployment Value:\\
Decision-Aligned Reliability Evaluation for\\
Foundation MLIPs in Materials Discovery}

\author{%
  Harry Sevi \\
  Altrove, Paris, France \\
  \texttt{harry.sevi@altrove.ai} \\
  \And
  Joonatan E.~M.~Laulainen \\
  Altrove, Paris, France \\
  \texttt{joonatan.laulainen@altrove.ai} \\
}

\begin{document}
\maketitle

\begin{abstract}
In AI-guided materials discovery, machine-learning models
increasingly decide which candidate materials should receive
expensive first-principles or experimental evaluation. Standard
reliability metrics such as uncertainty, calibration, novelty,
and ranking accuracy evaluate properties of the
\emph{predictions}, but success on these metrics does not
necessarily imply a better downstream screening decision. We
study this mismatch for foundation machine-learned interatomic
potentials (MACE-MP, CHGNet, Orb~V3) used to rank candidate
crystal structures before DFT verification. We evaluate three
representative reliability interventions (novelty detection,
sparse DFT calibration, and chemistry-level ranking audits) using
a deployment objective that measures how much DFT can be avoided
while maintaining a required recall of near-hull candidates. Across all three, a reliability signal can succeed on its
immediate target without improving the final screening decision:
novelty detects distribution shift but weakly ranks failure
severity; calibration can improve pooled ranking metrics while
leaving screening unchanged or degrading it; and sparse audits
identify MLIP--chemistry pairs with poor ranking reliability
without yielding a robustly beneficial routing policy. A two-oracle analysis further shows that the source of failure
depends on the screening decision itself: under formation-energy
screening, even perfect knowledge of average ranking reliability can select the
wrong model, whereas in the evaluated hull-aligned variant the
oracle target becomes decision-aligned but finite-budget audits
still fail to realise the value. Reliability should be evaluated
relative to the decision it supports: the relevant question is
not whether a model is uncertain or inaccurate, but whether
acquiring and acting on additional information will improve the
final discovery decision enough to justify its cost.
\end{abstract}

\section{Reliability is a decision problem}
\label{sec:intro}

In materials-discovery pipelines, foundation MLIPs such as
MACE-MP~\cite{batatia2022mace}, CHGNet~\cite{deng2023chgnet}, and
Orb~V3~\cite{rhodes2025orb} increasingly
prefilter~\cite{creed2026six,wehrhan2025mlipaudit} large candidate
pools~\cite{merchant2023scaling,zeni2025generative} before expensive
first-principles evaluation. Most
reliability work asks whether the model is uncertain,
out-of-distribution, calibrated, or rank-consistent. An automated
laboratory, however, needs a different answer: \emph{will acquiring and
acting on this reliability information save high-fidelity evaluation
without losing promising materials?} A model can look reliable by
standard predictive metrics and still induce a poor screening decision, because
these metrics score predictions rather than the downstream action they
trigger.

We make this operational. We group candidates by chemical system
(\emph{chemsys}): structures containing the same set of elements,
e.g.\ Mn--Na--O. Within each chemsys, an MLIP ranks candidates by
predicted formation energy and the top fraction $f$ is sent to DFT; a
near-hull candidate has
$E^{\mathrm{DFT}}_{\mathrm{hull}}\!\le\!25$~meV/atom.\footnote{A
prespecified operational threshold within the tens-of-meV/atom
metastability scale of known inorganic
crystals~\cite{sun2016thermodynamic}, not a universal synthesizability
criterion; the framework applies to any $\tau_{\rm hull}$.} The
deployment utility is the DFT saved at a required near-hull recall $R$,
\[
S_R \;=\; 1-\frac{\text{unique DFT calls to reach recall }R}{N},
\]
so larger $S_R$ means fewer DFT calculations at the same recall. The
decision-relevant question is whether a reliability intervention
improves $S_R$ enough to offset the DFT it spends, its net value of
sample information (App.~\ref{app:pool}).

\begin{figure}[t]
\centering
\resizebox{\textwidth}{!}{%
\begin{tikzpicture}[
  font=\footnotesize,
  every node/.style={align=center},
  chip/.style={draw,rounded corners=2pt,inner sep=4pt,align=center,
               minimum width=3.3cm},
  head/.style={chip,fill=blue!8,font=\footnotesize\bfseries,minimum height=0.9cm},
  good/.style={chip,fill=green!10,minimum height=1.1cm},
  bad/.style={chip,fill=red!8,minimum height=1.1cm},
  rowlab/.style={font=\scriptsize\itshape,text=gray!45!black,align=right,text width=1.8cm},
  arr/.style={->,>={Latex[length=3.5pt]},shorten >=1pt,shorten <=1pt,thick,gray!60!black},
]
\node[head] (novH) at (0,0)   {Novelty score};
\node[head] (calH) at (5,0)   {Sparse DFT calibration\\[1pt]{\scriptsize\mdseries\color{gray!55!black}$K\!=\!10$ canonical setting}};
\node[head] (audH) at (10,0)  {Chemistry-level DFT audit\\[1pt]{\scriptsize\mdseries\color{gray!55!black}$K\!=\!10$ canonical setting}};
\node[good] (novG) [below=6mm of novH] {$\checkmark$~detects\\ distribution shift\\[1pt]{\scriptsize\color{gray!55!black}AUROC $0.995$--$1.000$}};
\node[good] (calG) [below=6mm of calH] {$\checkmark$~validation\\ metric improves\\[1pt]{\scriptsize\color{gray!55!black}pairwise $+1.4$ to $+3.5$~pp}};
\node[good] (audG) [below=6mm of audH] {$\checkmark$~unreliable MLIP--chemistry\\ pairs detected\\[1pt]{\scriptsize\color{gray!55!black}TPR $\!\sim\!0.85$ @ FPR $\!\sim\!0.9\%$}};
\node[bad] (novB) [below=6mm of novG] {$\times$~poorly ranks\\ failure severity\\[1pt]{\scriptsize\color{gray!55!black}$|\rho|<0.35$}};
\node[bad] (calB) [below=6mm of calG] {$\times$~screening\\ does not improve\\[1pt]{\scriptsize\color{gray!55!black}$\Delta S_R\!\approx\!0$ or negative}};
\node[bad] (audB) [below=6mm of audG] {$\times$~routing does not\\ robustly save DFT\\[1pt]{\scriptsize\color{gray!55!black}reliable $\Delta S_R\!<\!0$}};
\node[rowlab] (rl1) [left=3mm of novH] {Reliability method};
\node[rowlab] (rl2) [left=3mm of novG] {Immediate target};
\node[rowlab] (rl3) [left=3mm of novB] {Downstream decision};
\draw[arr] (novH) -- (novG); \draw[arr] (novG) -- (novB);
\draw[arr] (calH) -- (calG); \draw[arr] (calG) -- (calB);
\draw[arr] (audH) -- (audG); \draw[arr] (audG) -- (audB);
\node[draw,rounded corners=3pt,fill=blue!5,font=\bfseries\footnotesize,
      minimum width=12.5cm,align=center,inner sep=5pt]
      at (5,-5.5)
      {Diagnostic success does not guarantee deployment value};
\end{tikzpicture}%
}
\caption{Three reliability strategies, each hitting an immediate target
(green, $\checkmark$) but not the next decision-relevant one (red,
$\times$): novelty detects shift but weakly ranks per-structure failure
severity; calibration improves pooled ranking but not within-chemsys
screening; the audit flags unreliable MLIP--chemistry pairs but
its routing rule does not robustly save DFT. At chemistry level, direct
aggregation of the continuous novelty score enriches early audit
allocation but remains incomplete (App.~\ref{app:continuous-novelty}).
Numbers in \S\ref{sec:findings}; $K\!=\!10$ is a canonical operating
point (App.~\ref{app:scoring-note}). Calibration $\Delta S_R$ excludes
fitting-label cost to isolate score correction; audit $\Delta S_R$
charges the audit labels.}
\label{fig:three-uses}
\end{figure}
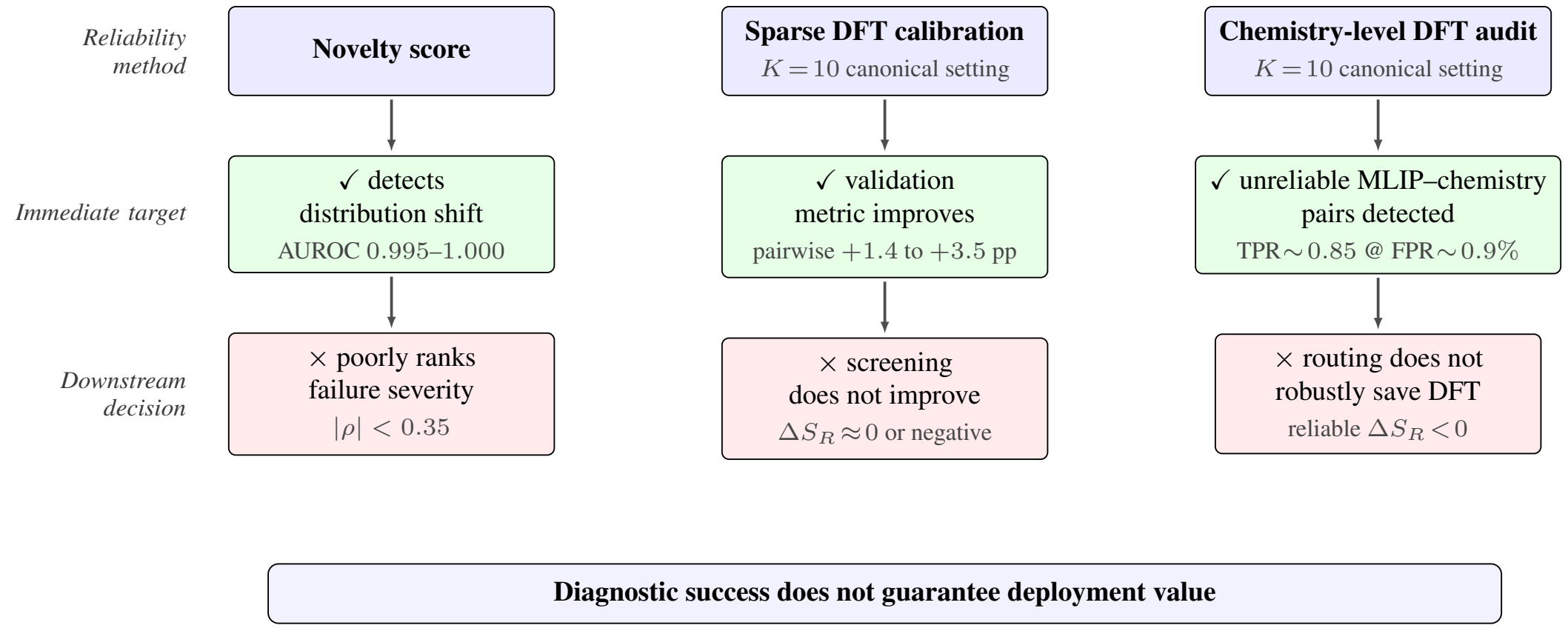

Using three representative interventions (a zero-DFT novelty
diagnostic, sparse DFT calibration, and a chemistry-level ranking
audit) as case studies, we expose an overlooked evaluation problem:
\emph{success on a reliability signal's native objective does not
guarantee deployment value}, and it can fail for structurally different
reasons (Fig.~\ref{fig:three-uses}). An oracle analysis then locates the
failure: under formation-energy screening the reliability target is
\emph{misaligned} with near-hull discovery; after aligning the ranked
quantity with predicted hull distance the target becomes
decision-relevant, but finite-budget acquisition still faces an
\emph{accuracy--cost trade-off}.

\paragraph{Contributions.} We make three contributions. First, we place
novelty detection, calibration, and ranking audits in a common screening
protocol and evaluate each by the DFT it saves at a fixed near-hull
recall, not by predictive or diagnostic metrics alone. Second, we show
that each intervention can improve its native reliability metric without
improving the screening decision, in three structurally different ways.
Third, oracle experiments identify two distinct reasons for this gap: the
reliability target can be misaligned with near-hull discovery, or, once
aligned, the information required to act on it can cost more than the
decision improvement it enables (\S\ref{sec:selection}).

\paragraph{Related work.} Foundation MLIPs are compared by predictive
error and ranking metrics; Matbench Discovery~\cite{riebesell2025framework}
scores them as prefilters with recall, precision, and
discovery-acceleration. Structure-based OOD
benchmarks~\cite{omee2024structure}, softening
diagnostics~\cite{deng2025systematic}, MatUQ~\cite{tan2026matuq},
MLIPAudit~\cite{wehrhan2025mlipaudit}, and calibrated MLIP
uncertainty~\cite{ho2026flexible} show predictive accuracy and
uncertainty quality decouple; \emph{Proof-Carrying Materials}~\cite{basu2026proof} is closest in
spirit: it audits MLIP-based stability-filter decisions and learns a risk
model for when those decisions may fail. These evaluate model or
diagnostic quality, not
reliability interventions as costly information-acquisition steps.
Decision-focused learning~\cite{elmachtoub2022smart,wilder2019melding}
and decision-aligned UQ~\cite{schneider2026decision} judge predictions by
the actions they induce; active and multi-fidelity
learning~\cite{settles2009active,kang2025accelerating,peherstorfer2018survey}
choose queries to improve the model. To our knowledge no prior work
evaluates multiple MLIP reliability interventions as information-acquisition
policies under a common fixed-recall DFT-saving objective.

\section{Decision-aligned screening evaluation}
\label{sec:benchmark}

Every intervention is evaluated inside the same downstream decision. At
retain fraction $f$, each chemsys $c$ sends its top $\lceil f
N_c\rceil$ MLIP-ranked candidates to DFT. Let $C_\pi(f)$ be the number of
\emph{unique} DFT calculations a policy $\pi$ requires at that retain
fraction, including any information-acquisition calls charged to $\pi$,
and let $f_R^\star\!=\!\min\{f:\text{near-hull recall}(f)\!\ge\!R\}$ be the
smallest retain fraction at which the retained candidates recover at least
a fraction $R$ of all DFT near-hull structures. The deployment saving is
\[
S_R(\pi) \;=\; 1-\frac{C_\pi(f_R^\star)}{N},
\]
with $N$ the total pool size, so $S_R\!=\!0.70$ means the policy reaches
the required recall while avoiding $70\%$ of the DFT calculations. $S_R$
is an \emph{offline cost--recall frontier}: $f_R^\star$ is chosen
retrospectively from complete DFT labels to compare policies at matched
recall, not a prospective stopping rule a laboratory could apply without
the labels. Because a $K$-sample audit is random, we report both
\emph{expected} recall (met on average over $200$ draws) and
\emph{reliable} recall (met on $\ge\!95\%$ of draws). Full accounting and
the value-of-information formalism are in App.~\ref{app:pool}.

We evaluate three foundation MLIPs (MACE-MP-medium and CHGNet-v0.3.0,
trained on MPtrj; Orb~V3, OMat24). Our main \emph{deployment} benchmark is
a $15$-chemsys Materials Project~\cite{jain2013commentary} subset
($N_c\!\ge\!12$; $1{,}147$ structures). GNoME
quaternaries~\cite{merchant2023scaling} and a Zintl-phase set (assembled
from Materials Project Zintl-phase chemistries;
App.~\ref{app:continuous-novelty}) are used only as shifted
\emph{stress-test} pools for the novelty and audit diagnostics; they are
not part of the deployment benchmark. A separate, disjoint $250$-chemsys
Materials Project \emph{robustness} pool provides preregistered
out-of-fold policy evaluation (App.~\ref{app:gate2}). The three interventions span
zero-cost and $K\!=\!10$ DFT settings: a zero-DFT novelty score,
$K$-sample calibration of MLIP energies, and a $K$-sample ranking audit.
We use $K\!=\!10$ as a canonical operating point, with budget
sensitivity in App.~\ref{app:scoring-note}. Calibration $\Delta S_R$ excludes the $K$
fitting labels to isolate whether score correction itself helps the
decision; audit $\Delta S_R$ charges every DFT call.

\section{Three proxy-to-decision gaps}
\label{sec:findings}

\subsection{Novelty detects shift but only partially localizes failure}
\label{sec:novelty}
A Mahalanobis\textsuperscript{++} novelty
score~\cite{muller2025mahalanobis++} over each MLIP's per-atom embeddings
(model-specific training-side references) raises three questions. First,
can it detect distribution shift? It separates $1{,}225$ MP structures
from $300$ GNoME quaternaries almost perfectly: area under the ROC curve
(AUROC) $=\!0.995/0.998/1.000$ on MACE-MP / Orb~V3 / CHGNet. Second, do
more novel structures have larger MLIP errors? Not reliably: the Spearman
correlation between the score and absolute formation-energy error is
$+0.32/-0.10/+0.16$ (all $|\rho|\!<\!0.35$, sign not preserved), and
re-estimating Orb~V3's reference leaves this severity correlation
essentially unchanged (App.~\ref{app:continuous-novelty}). Third, can it
decide where to spend DFT? Aggregating the continuous score to the chemsys
level does enrich early discovery of failure cases (mean and
$90$th-percentile pooling recover $3$ of the $5$ MLIP--chemistry pairs the
audit later flags as unreliable (\S\ref{sec:audit}), by $M\!=\!10$ audited
chemsys vs.\ $1.42$ at random), but it misses an
in-distribution failure (Mn--Na--O ranks $19$--$24$ of $35$). Novelty is
therefore useful for detecting shift but incomplete as a guide to
deployment failure.

\subsection{Calibration can improve validation without improving the decision}
\label{sec:calibration}
We spend the $K\!=\!10$ DFT budget to correct MLIP formation energies per
chemsys and ask whether a standard validation gain transfers to screening
(Table~\ref{tab:calibration}). As the validation metric we use
\emph{pooled pairwise accuracy}: the fraction of held-out candidate pairs
whose predicted energy ordering agrees with DFT, pooling pairs across all
chemistries. A per-chemsys mean shift cannot change within-chemsys
ordering, so $\Delta S_R\!=\!0$ exactly while pooled pairwise improves
$+1.4/+1.5$~pp (it rewards the cross-chemsys comparisons the within-chemsys
policy never uses). An
unconstrained affine fit changes the decision only through rare
negative-slope inversions ($\sim\!2\%$ of fits), which are purely
harmful ($\Delta S_{0.95}\!=\!-3.7$~pp). A per-element correction can
genuinely reorder within-chemsys candidates and yields the largest
validation gain ($+3.2/+3.5$~pp), yet remains deployment-negative
($\Delta S_{0.95}\!=\!-3.3/-2.2$~pp). The qualitative mismatch also
persists under predicted-hull screening (App.~\ref{app:scoring-note}):
better validation does not imply a better decision.

\begin{table}[h]
\centering\small
\caption{Calibration: validation improves while deployment does not.
Pooled pairwise accuracy (validation) vs.\ $\Delta S_{0.95}$, the change
in DFT saved relative to the uncalibrated MLIP at $95\%$ near-hull recall
(positive means more DFT saved; MACE-MP / Orb~V3), $K\!=\!10$, fitting
labels not charged.}
\label{tab:calibration}
\begin{tabular}{@{}l l r@{}}
\toprule
\textbf{Correction} & \textbf{Pooled pairwise} & \textbf{$\Delta S_{0.95}$ (pp)} \\
\midrule
Per-chemsys mean shift & $+1.4/+1.5$~pp & $0.0 \;/\; 0.0$ \\
Unconstrained affine   & $+1.8/+1.8$~pp & $-3.7 \;/\; -3.7$ \\
Per-element (FERE-style)~\cite{stevanovic2012fere} & $+3.2/+3.5$~pp & $-3.3 \;/\; -2.2$ \\
\bottomrule
\end{tabular}
\end{table}

\subsection{The audit works diagnostically but not reliably operationally}
\label{sec:audit}
\label{sec:audit-diag}
The same $K\!=\!10$ budget can instead \emph{measure} within-chemsys
ranking reliability, leaving MLIP scores untouched. For each MLIP $m$ and
chemsys $c$ we consider the \emph{MLIP--chemistry pair} $(m,c)$, and call
it \textsc{unsafe} when the MLIP correctly orders fewer than $75\%$ of all
candidate pairs within that chemistry (population pairwise accuracy
$A^{\text{pop}}_{m,c}\!<\!0.75$); a $K\!=\!10$ audit estimates this from
the $\binom{K}{2}\!=\!45$ pairwise comparisons among ten randomly sampled
DFT-labelled structures. Across $105$ MLIP--chemistry pairs (3 MLIPs
$\times$ 35 chemsys from MP $\cup$ Zintl $\cup$ GNoME) with five
population-unsafe pairs, one-shot audits identify unsafe pairs with
true-positive rate (TPR)~$\approx\!0.85$ at false-positive rate
(FPR)~$\approx\!0.009$, and averaging $30$ draws orders all five unsafe
pairs below the $100$ safe ones; the signal is informative.
(These numbers characterise sampling variability conditional on this
benchmark's five positives, not unsafe-case prevalence.)

\label{sec:audit-routing}
\label{sec:decomp}
Acting on the verdict is another matter. The natural rule
(\textsc{unsafe}$\to$all-DFT, else MLIP top-$f$ verify) faces an economic
problem: the audit pays $K$ DFT calculations on \emph{every} chemistry,
while its benefit is realised only on the rare chemistries where the MLIP
is genuinely unsafe (one of $15$ MP chemsys). So the audit is net-negative
at $R\!=\!0.90$ under either recall criterion
(Table~\ref{tab:audit-deployment}); at $R\!=\!0.95$ larger tail headroom
turns the \emph{expected}-recall value positive on MACE-MP/Orb~V3 (the
Mn--Na--O rescue), but the one-shot audit misses the unsafe cell in
$\sim\!15\%$ of draws, so the gain vanishes under the \emph{reliable}-recall
requirement. Both the $K$ measurements and the acted-on verdict contribute
to the loss (decomposition in App.~\ref{app:audit-robustness}): the signal
is useful, but the action costs DFT everywhere while benefit exists only on
rare chemistries.

\begin{table}[h]
\centering\small
\caption{\textbf{Net deployment value of audit routing},
$\Delta S_R\!=\!S_R(\text{audit})-S_R(\text{MLIP-only})$ (pp); negative
means it costs more DFT than it saves. $200$ seeds.}
\label{tab:audit-deployment}
\begin{tabular}{cl rrr}
\toprule
\textbf{Recall} & \textbf{Criterion} & \textbf{MACE-MP} & \textbf{Orb~V3} & \textbf{CHGNet} \\
\midrule
$R\!=\!0.90$ & Expected                 & $-3.3$ & $-4.1$ & $-3.8$ \\
             & Reliable ($\ge\!95\%$)   & $-6.4$ & $-6.4$ & $-4.8$ \\
$R\!=\!0.95$ & Expected                 & $+3.1$ & $+1.6$ & $-0.8$ \\
             & Reliable ($\ge\!95\%$)   & $-2.0$ & $-4.5$ & $-1.6$ \\
\bottomrule
\end{tabular}
\end{table}

\section{Why does translation fail?}
\label{sec:selection}

The previous sections show three proxy-to-decision gaps; oracle analysis
distinguishes why they arise. Is the reliability signal measuring the
wrong property, or is the right property too costly to estimate? We
separate these explanations with chemistry-specific model selection.
Instead of routing an unreliable chemistry to all-DFT, we reuse the
$K\!=\!10$ audit to choose which of the three MLIPs should rank that
chemistry; the baseline uses the single best-fixed MLIP (Orb~V3) for every
chemistry. Audit-based selection loses ($-5.1/-4.8$~pp at
$R\!=\!0.90/0.95$), so we introduce two free oracles to diagnose why
(Table~\ref{tab:selection}). A \emph{pairwise oracle} removes audit
sampling noise: for each chemistry it picks the MLIP with the best true
formation-energy ranking accuracy $A^{\text{pop}}_{m,c}$. A
\emph{deployment oracle} instead picks the MLIP with the best near-hull
screening outcome. Comparing them asks whether ranking reliability is
aligned with the deployment objective.

\paragraph{Target misalignment (formation-energy regime).} Under
formation-energy screening the pairwise oracle \emph{still} loses
($-3.7$~pp), whereas the deployment oracle \emph{beats} best-fixed
($+0.9/+2.9$~pp):
$S_{\text{pair-oracle}}\!<\!S_{\text{best-fixed}}\!<\!S_{\text{deploy-oracle}}$.
Useful chemistry-specific model choice therefore exists, but average
formation-energy ranking accuracy does not identify it: pairwise
accuracy weights all inversions equally, whereas high-recall screening
depends on the low-energy tail. The reliability target is misaligned with
the decision.

\paragraph{Accuracy--cost trade-off (hull-aligned regime).} We then change
the ranked quantity from formation energy to predicted distance above the
convex hull (MACE-MP and CHGNet; App.~\ref{app:scoring-note}). The
pairwise oracle now flips to $+1.1/+4.1$~pp and at $R\!=\!0.95$ matches the
deployment oracle: the target is aligned. Yet the finite-budget audit
still loses ($-5.0/-1.8$~pp): as the audit grows, selector accuracy
improves but net deployment value falls (better diagnosis, worse
decision), because small audits are noisy and larger ones consume the
gain. A
disjoint $250$-chemsys out-of-fold test reproduces the pattern: a cheap
disagreement signal predicts chemistry-conditional headroom
($\rho\!=\!+0.22$) yet the composed policy is deployment-neutral ($\Delta
S_{0.95}\!=\!-0.04$~pp, CI $[-1.72,+2.47]$; App.~\ref{app:gate2}).

\begin{table}[h]
\centering\small
\caption{\textbf{Why reliability fails to create value}, expected-recall
$\Delta S_R$ vs.\ best-fixed MLIP under formation-energy screening. The
pairwise oracle isolates whether removing audit noise suffices; the
deployment oracle, whether aligned selection could help. Under
predicted-hull scoring the pairwise-oracle sign flips positive
(App.~\ref{app:scoring-note}).}
\label{tab:selection}
\begin{tabular}{@{}l rr@{}}
\toprule
\textbf{Policy (model choice per chemsys)} & $\Delta S_{0.90}$ & $\Delta S_{0.95}$ \\
\midrule
Samples-only ($K$ labels, no switch)   & $-1.8$ & $-2.3$ \\
Audit-select (estimated pairwise)      & $-5.1$ & $-4.8$ \\
Pairwise oracle (true pairwise acc.)   & $-3.7$ & $-3.7$ \\
Deployment oracle (true near-hull quality) & $+0.9$ & $+2.9$ \\
\bottomrule
\end{tabular}
\end{table}

\section{Open challenge: optimise decision value, not proxy accuracy}
\label{sec:discussion}

MLIP reliability research largely asks whether a diagnostic predicts
error, detects OOD inputs, or improves calibration. Our case studies show
these are insufficient deployment objectives: the same signal can succeed
natively yet fail, or even harm, the screening decision. The central open
challenge is therefore how an automated materials-discovery system should
decide \emph{whether, where, and how much} high-fidelity information to
acquire: choosing measurements by their \emph{expected improvement in the
downstream decision, minus their measurement cost} (formally, a net value
of sample information; App.~\ref{app:pool}), not by reliability accuracy.
We see three directions. \textbf{(1)~Decision-aligned targets:} measure reliability for
the part of the ranking that actually determines discovery (the
low-energy / hull tail rather than global ranking), since target
alignment alone removed the misalignment failure here.
\textbf{(2)~Adaptive acquisition:} do not fix $K$ universally; acquire DFT
only where its expected decision benefit exceeds its cost, the quantity
our accuracy--cost frontier makes explicit. \textbf{(3)~Prospective
evaluation:} evaluate sequential policies that choose what to measure
without access to unrevealed DFT labels, not only retrospective
diagnostics or the offline frontier $S_R$ used here.

\paragraph{Limitations.} Evaluation is offline and retrospective on a
small deployment pool ($15$ chemsys) and the specific policies tested;
the disjoint $250$-chemsys extension is deployment-neutral, and all
conclusions are regime-conditional (formation-energy vs.\ predicted-hull
scoring).

\paragraph{Conclusion.}\label{sec:conclusion}
Successful reliability diagnostics do not automatically improve materials
screening. In our experiments they fail for two reasons: under
formation-energy screening the diagnostic target is misaligned with
near-hull discovery; once aligned with the hull objective, finite audits
are limited by an information-quality/DFT-cost trade-off. Reliability for
automated discovery should therefore be evaluated, and acquired, by the
downstream DFT it saves at the required recall, not by proxy accuracy
alone.

\bibliographystyle{plain}
\bibliography{ai4mat}

\newpage
\appendix

\noindent Table~\ref{tab:appendix-overview} maps every robustness check in
this appendix to the question it probes and its verdict for the paper's
central finding, that diagnostic success does not imply deployment value.
The sections that follow give the full analyses.

\begin{table}[ht]
\centering\small
\caption{\textbf{Appendix overview: robustness checks and verdicts.} Each
check stresses the central finding that diagnostic success does not imply
deployment value: \emph{holds} = the finding survives; \emph{qualified} =
survives with the stated caveat; \emph{context} = establishes a baseline
rather than testing the claim. The \emph{Check} column points to the
subsection (and its data table).}
\label{tab:appendix-overview}
\setlength{\tabcolsep}{4pt}
\begin{tabular}{@{}l >{\raggedright\arraybackslash}p{3.9cm} >{\raggedright\arraybackslash}p{4.7cm} l@{}}
\toprule
Check & Probes & Result & Verdict \\
\midrule
\ref{app:scoring-note}\,(i)    & formation-energy vs.\ predicted-hull baseline & Hull baseline stronger ($+8$--$19$~pp); formation-energy is the conservative choice & context \\
\ref{app:scoring-note}\,(ii)   & audit routing under hull scoring              & Audit-tax strengthens at $R\!=\!0.90$; the one positive cell flips negative & holds \\
\ref{app:scoring-note}\,(iii)  & calibration under hull scoring                & Mean-shift / per-element / affine ($a\!>\!0$) exactly invariant & holds \\
\ref{app:scoring-note}\,(iv)   & model selection under hull scoring            & Free oracle target-aligns, yet the finite audit still loses & holds \\
\ref{app:scoring-note}\,(v)    & audit-select over $K\!\in\!\{2,\dots,10\}$    & Negative at every $K$: an accuracy--cost trade-off & holds \\
\ref{app:scoring-note}\,(vi)   & is the oracle headroom proxy slack?           & No: not explained by formation-to-hull rank distortion & holds \\
\ref{app:scoring-note}\,(vii)  & macro vs.\ micro $S_R$ weighting              & Survives; individual routing/oracle cells are weighting-sensitive & qualified \\
\ref{app:scoring-note}\,(viii) & $2\!\times\!2$ representation $\times$ model error & Representation dominates; the two interact & holds \\
\ref{app:scoring-note}\,(ix)   & normalised audit-budget sweep                 & Gap not specific to $K\!=\!10$: accuracy rises as value falls & holds \\
\midrule
App.~\ref{app:calibration}      & three calibration families, two ranking rules & Deployment-neutral or negative throughout & holds \\
App.~\ref{app:audit-robustness} & prevalence, continuous novelty, relaxation    & Value regime-dependent; novelty partial; robust to relaxation & qualified \\
App.~\ref{app:tail}             & tail-aligned audit selectors                  & Move toward the oracle, none deployment-positive & holds \\
App.~\ref{app:gate2}            & sequential gating at $250$-chemsys scale      & Signal exists and predicts, yet deployment-neutral & holds \\
\bottomrule
\end{tabular}
\end{table}

\section{Pool construction and DFT accounting}
\label{app:pool}

The MP reference contains $1{,}225$ entries across $25$ chemsys
that pass the $\text{deprecated}\!=\!\text{False}$ filter and for
which all three MLIPs returned a valid single-point energy. The deployment subset additionally requires
$N_c\!\geq\!K\!+\!2\!=\!12$ per chemsys, leaving $15$ chemsys with
$1{,}147$ structures (single-point) or $1{,}104$ structures
(MLIP-relaxed, using elemental references evaluated in the same
relaxation regime). Per-chemsys sizes and inclusion are listed in
Table~\ref{tab:app-mp-chemsys}. Across the same $25$ chemsys,
$1{,}178$ MLIP-relaxed geometries completed for each MLIP without
error; $92$--$93\%$ additionally satisfied the FIRE convergence
criterion $|f|_{\max}\!<\!0.05$~eV/{\AA} on all atoms. Unique-candidate DFT accounting charges
each material at most once across the audit sample $A_s$ and
downstream verify set $T_f(s)$; this alone removes roughly $150$
double-counted DFT calls per seed and materially changes the
audit-routing $\Delta S_R$ at $R\!=\!0.90$.

Formally, each reliability intervention is an information-acquisition
action $I$ with DFT cost $c(I)$, and its net value for the screening
decision $D$ is the expected value of sample information minus that cost,
$\mathrm{EVSI}(I\!\mid\!D)-c(I)$; the deployment saving $S_R$
operationalises this as unique DFT saved at fixed near-hull recall, so
$\Delta S_R$ is the empirical net value of information of an intervention.

\begin{table}[h]
\centering\footnotesize
\caption{Per-chemsys Materials Project pool composition. $N_c$
counts single-point-successful entries. Deployment inclusion
requires $N_c\!\ge\!12$; excluded chemsys are used only for the
zero-cost novelty diagnostic in \S\ref{sec:novelty}. Ordered by
$N_c$ (descending).}
\label{tab:app-mp-chemsys}
\begin{tabular}{@{}l r c@{\hskip 1.5em}l r c@{}}
\toprule
Chemsys & $N_c$ & Deploy? & Chemsys & $N_c$ & Deploy? \\
\midrule
O--Si     & 337 & Y & Al--Ni      & 13 & Y \\
Fe--Li--O & 176 & Y & Cl--Na      & 13 & Y \\
Fe--O     & 140 & Y & H--Ti       & 12 & Y \\
O--Ti     & 124 & Y & Li--S       & 11 & \\
Mn--Na--O &  93 & Y & H--Mg       & 10 & \\
Ba--O--Ti &  53 & Y & C--Ti       &  9 & \\
Ca--O--Ti &  45 & Y & Br--Cs--Pb  &  8 & \\
Ga--N     &  44 & Y & Ca--F       &  8 & \\
Fe--S     &  35 & Y & Li--P--S    &  8 & \\
Mg--O     &  28 & Y & B--Ti       &  7 & \\
Ni--Ti    &  18 & Y & Bi--Cl--O   &  7 & \\
N--Ti     &  16 & Y & N--O--Ta    &  6 & \\
          &     &   & Al--N       &  4 & \\
\bottomrule
\end{tabular}
\end{table}

\subsection{Decision-representation and policy robustness}
\label{app:scoring-note}
Section~\ref{sec:benchmark} ranks candidates within each chemsys by
MLIP-predicted formation energy per atom, matching the deployment
convention of applying the MLIP as a fast formation-energy prefilter
prior to the DFT stability check. We evaluate a sequence of robustness
questions about this choice, each mapped to its verdict in
Table~\ref{tab:appendix-overview}.\footnote{Orb~V3 is excluded from this
appendix: its hull-support predictions are not in our release, so no
predicted $E_{\text{above-hull}}$ is available. The check is limited
to the two MLIPs with a materialised MLIP-hull.}

\paragraph{(i) The MLIP-only baseline is scoring-rule sensitive.}
Replacing the formation-energy prefilter with an MLIP-predicted
$E_{\text{above-hull}}$ (constructed from the same MLIP formation
energies over the per-chemsys MP support entries) raises MLIP-only
$S_R$ by $+8.4/+15.0$~pp on MACE-MP and by $+11.4/+19.4$~pp on CHGNet
at $R\!\in\!\{0.90,0.95\}$. For the two MLIPs evaluated, the
formation-energy baseline is therefore more conservative than
the predicted-hull baseline.

\paragraph{(ii) The audit-policy delta keeps its sign at $R\!=\!0.90$
but flips at $R\!=\!0.95$ (mean-recall).}
We reran the audit-routing simulator with two coordinated changes:
the deployment score \emph{and} the audit's within-chemsys
pairwise-accuracy diagnostic both use MLIP-predicted
$E_{\text{above-hull}}$ (vs.\ DFT-computed $E_{\text{above-hull}}$
for the audit). All other protocol details ($K\!=\!10$, $\tau_{\rm unsafe}
= 0.75$, $\tau_{\rm safe} = 0.85$, $200$ seeds, dense retain grid,
unique-DFT accounting) are unchanged. As a sanity check the same
simulator on the formation-energy score reproduces Table~\ref{tab:audit-deployment}
within rounding (MACE-MP $-3.3$, Orb~V3 $-4.1$, CHGNet $-3.8$~pp at
$R\!=\!0.90$ expected-recall).
Predicted-hull screening raises the MLIP-only baseline, reducing
the headroom available to justify a fixed-cost audit;
Table~\ref{tab:audit-ehull} reports the resulting policy deltas.

\begin{table}[!ht]
\centering\small
\caption{$\Delta S_R = S_R(\text{audit}) - S_R(\text{MLIP-only})$
(percentage points, mean-recall) under the two scoring rules. The
$R\!=\!0.90$ audit deficit is $2$--$4\times$ larger under
$E_{\text{above-hull}}$; the $R\!=\!0.95$ MACE-MP mean-recall
carveout (+3.2 pp under $E_{\text{form}}$) flips negative. Orb~V3
is not evaluable in the predicted-hull column.}
\label{tab:audit-ehull}
\begin{tabular}{c l rr}
\toprule
$R$ & MLIP & $\Delta$ ($E_{\text{form}}$, main body) & $\Delta$ ($E_{\text{above-hull}}$, this check) \\
\midrule
$0.90$ & MACE-MP & $-3.2$ & $\mathbf{-14.1}$ \\
       & CHGNet  & $-3.6$ & $\mathbf{-7.5}$ \\
\midrule
$0.95$ & MACE-MP & $+3.2$ & $\mathbf{-9.1}$ (sign flip) \\
       & CHGNet  & $-0.8$ & $\mathbf{-6.7}$ \\
\bottomrule
\end{tabular}
\end{table}
\FloatBarrier

At $R\!=\!0.90$, the audit-tax conclusion persists and strengthens
under predicted-hull scoring. This is consistent with reduced
headroom, although the audit diagnostic also changes under the
predicted-hull variant. Writing the net value of the routing rule as
$\Delta S_R = H_R - T_R$ (headroom $H_R$, realised only on the rare
unsafe chemsys, minus the routing tax $T_R$, the $K$-DFT audit paid on
every chemsys), the tax $T_R$ is not held constant across scoring rules,
so we do not attribute the full change in $\Delta S_R$ to headroom alone. At $R\!=\!0.95$ the same mechanism
explains the sign flip: under the weaker formation-energy baseline
the Mn--Na--O rescue produced enough apparent headroom to give
$+3.2$~pp on MACE-MP; under the stronger predicted-hull baseline
that headroom no longer pays for the audit, and the result
becomes $-9.1$~pp. The single positive cell in
Table~\ref{tab:audit-deployment} is therefore scoring-rule
fragile; the audit-tax conclusion is not.

\paragraph{(iii) Calibration invariance under
$E_{\text{above-hull}}$.} Applied consistently to both the candidate
and its hull-reference predictions, a per-chemsys mean-shift $\Delta_c$
cancels pointwise:
\[
E_{\rm hull}^{\prime}(x_i) \;=\;
(E_{\rm form}(x_i)\!+\!\Delta_c) - (h^{\rm ref}(x_i)\!+\!\Delta_c)
\;=\; E_{\rm form}(x_i) - h^{\rm ref}(x_i) \;=\; E_{\rm hull}(x_i),
\]
so ranking and $S_R$ are exactly unchanged. The implementation in
\S\ref{app:calibration} applies $\Delta_c$ to candidate formation
energies only, which acts as the same uniform within-chemsys shift on
$E_{\rm hull}$ and therefore leaves per-chemsys top-$f$ ranking invariant. Either way,
$\Delta S_R^{\rm mean\_shift}\!=\!0$ under $E_{\text{above-hull}}$
scoring by construction; the mean-shift falsification survives the
scoring-rule swap without further simulation.
The identity extends to consistently-applied per-element and
affine corrections. Under per-element $E_{\rm form}^{\prime}(x_i)
= E_{\rm form}(x_i) + \sum_e \delta_e\, f_e(x_i)$, the hull line at
composition $c$ is a convex combination
$h^{\rm ref}(c) = \min_w \sum_i w_i E_{\rm form}(x_i)$ over support
entries with $\sum_i w_i \mathbf{c}_i = \mathbf{c}$;
composition-linearity gives $h^{\prime\,\rm ref}(c) = h^{\rm ref}(c)
+ \sum_e \delta_e f_e(c)$, so
$E_{\rm hull}^{\prime}(x_i) = E_{\rm hull}(x_i)$. Under affine
$E_{\rm form}^{\prime} = a\,E_{\rm form} + b$ with $a > 0$,
$h^{\prime\,\rm ref}(c) = a\,h^{\rm ref}(c) + b$ so
$E_{\rm hull}^{\prime} = a\,E_{\rm hull}$ and ranking is invariant.
Consistently applied composition-linear calibration therefore has
\emph{zero incremental effect} on the hull-ranked screening
decision under mean-shift and per-element and for the $\ge\!97\%$
of affine fits with $a\!>\!0$ (App.~\ref{app:calibration} reports
$a\!<\!0$ on $2.3\%/1.9\%/0\%$ of MACE-MP/Orb~V3/CHGNet fits,
where the invariance argument does not apply). If the $K$ fitting
labels are charged, any net deployment value comes from the sampled
candidates themselves, not the calibration transformation; under
unique-candidate accounting this is the samples-only effect studied
in \S\ref{sec:decomp}.

\paragraph{(iv) Chemistry-conditional model selection under
$E_{\text{above-hull}}$.} \S\ref{sec:selection} shows that under
$E_{\rm form}$ scoring a pairwise-oracle chemistry-conditional
selector \emph{underperforms} the best fixed MLIP ($-3.7$~pp at both
$R\!=\!0.90$ and $R\!=\!0.95$) while a deployment oracle finds
positive headroom ($+0.9/+2.9$~pp), demonstrating a
signal-decision mismatch.
We reran the selection simulator (script matching \S\ref{sec:selection}'s
protocol, $K\!=\!10$, $200$ seeds, dense $f$-grid) under
$E_{\text{above-hull}}$ scoring on the MACE-MP + CHGNet subset. To
separate the scoring-rule effect from the MLIP-count effect we
also reran the paper's $E_{\rm form}$ protocol on the same $2$-MLIP
subset as a control.

\begin{table}[h]
\centering\small
\caption{Chemistry-conditional selection oracles across scoring
rules. $\Delta$ vs best-fixed MLIP, mean-recall $S_R$, percentage
points. MACE-MP + CHGNet only (Orb~V3 lacks hull-support
predictions).}
\label{tab:selection-ehull}
\begin{tabular}{c l rr}
\toprule
$R$ & Scoring rule & $\Delta$ pairwise-oracle & $\Delta$ deploy-oracle \\
\midrule
$0.90$ & $E_{\rm form}$ (2-MLIP control) & $-3.7$ & $+0.9$ \\
       & $E_{\text{above-hull}}$          & $\mathbf{+1.1}$ & $+1.6$ \\
\midrule
$0.95$ & $E_{\rm form}$ (2-MLIP control) & $-2.5$ & $+0.0$ \\
       & $E_{\text{above-hull}}$          & $\mathbf{+4.1}$ & $+4.1$ \\
\bottomrule
\end{tabular}
\end{table}

The $E_{\rm form}$ 2-MLIP control replicates the paper's
signal-decision mismatch ($-3.7/-2.5$~pp), so the effect is not an
artefact of the MLIP-count reduction. Under $E_{\text{above-hull}}$
scoring the pattern flips: the pairwise-hull oracle
\emph{outperforms} best-fixed by $+1.1/+4.1$~pp, and at $R\!=\!0.95$
matches the deployment-oracle exactly ($+4.1$ vs $+4.1$). Population
pairwise accuracy on predicted-vs-DFT $E_{\text{above-hull}}$ is a
decision-aligned selector on this pool; population pairwise on
formation energies is not. The paper's ``pairwise oracle fails''
finding is therefore \emph{scoring-rule specific}: it belongs to the
formation-energy regime and does not survive alignment of the ranked
quantity with the deployment target.

Two other elements of \S\ref{sec:selection} are scoring-rule
robust. First, chemistry-conditional headroom exists in both regimes;
it is at least as large under $E_{\text{above-hull}}$ ($+4.1$~pp at
$R\!=\!0.95$) as under $E_{\rm form}$. Second, the audit-select policy still loses in the aligned regime
($-5.0$~pp at $R\!=\!0.90$; $-1.8$~pp at $R\!=\!0.95$). The residual
failure is no longer target misalignment: as the $K$-sweep below
shows, finite-sample selection error and the cost of acquiring the
audit jointly prevent the oracle value from being realised. At
$K\!=\!10, R\!=\!0.95$ even the cost-matched \emph{perfect} selector
is already $-1.0$~pp, so removing estimation error alone would not
turn the policy positive. The larger picture is unchanged:
improving the base decision representation delivers the
operational gain, whereas the downstream reliability intervention
does not.

\paragraph{(v) $K$-sweep of audit-select under $E_{\text{above-hull}}$.}
Since the pairwise-hull oracle is decision-aligned (iv), the
operational failure of audit-select under $E_{\text{above-hull}}$
($-5.0$/$-1.8$~pp at $R\!=\!0.90$/$0.95$) is a statement about
finite-sample estimation and audit cost, not about the target.
We swept the full feasible integer range,
$K\!\in\!\{2,\ldots,10\}$, under the $N_c\!\ge\!K\!+\!2$
eligibility rule on the fixed $15$-chemsys deployment pool
($200$ seeds; policies restricted to the four relevant here).
The cost-matched oracle always selects the population-best MLIP
for each chemistry but is charged the same $K$ DFT labels as
audit-select; \texttt{audit picks pop-best} reports the fraction
of audit decisions that select that same population-best MLIP.
Oracle-free is $K$-invariant at $+1.05/+4.10$~pp on this fixed
pool, as expected. Bold cells in Table~\ref{tab:ksweep} mark
positive cost-matched deltas (the accuracy-cost frontier at
$R\!=\!0.95$).

\begin{table}[h]
\centering\footnotesize
\setlength{\tabcolsep}{4pt}
\caption{Audit-select and oracle-select under $E_{\text{above-hull}}$
across the feasible $K$ range on the fixed $15$-chemsys pool.
$\Delta$ vs best-fixed MLIP, mean-recall percentage points.
Oracle-free ceiling is $K$-invariant at $+1.05/+4.10$~pp
($R\!=\!0.90/0.95$).}
\label{tab:ksweep}
\begin{tabular}{cc rrrrrrrrr}
\toprule
$R$ & Quantity & $K\!=\!2$ & $3$ & $4$ & $5$ & $6$ & $7$ & $8$ & $9$ & $10$ \\
\midrule
$0.90$ & audit-select $\Delta$        & $-0.8$ & $-1.0$ & $-1.7$ & $-1.9$ & $-2.6$ & $-3.3$ & $-3.5$ & $-4.3$ & $-5.0$ \\
       & oracle cost-matched $\Delta$ & $-0.3$ & $-0.5$ & $-1.2$ & $-1.9$ & $-2.6$ & $-3.4$ & $-3.6$ & $-4.3$ & $-5.0$ \\
       & audit picks pop-best         & $56\%$ & $59\%$ & $62\%$ & $65\%$ & $66\%$ & $68\%$ & $71\%$ & $73\%$ & $75\%$ \\
\midrule
$0.95$ & audit-select $\Delta$        & $-1.1$ & $-0.4$ & $-0.9$ & $-1.0$ & $-0.6$ & $-0.8$ & $-0.9$ & $-1.5$ & $-1.8$ \\
       & oracle cost-matched $\Delta$ & $\mathbf{+2.8}$ & $\mathbf{+2.6}$ & $\mathbf{+2.0}$ & $\mathbf{+1.4}$ & $\mathbf{+1.3}$ & $\mathbf{+0.6}$ & $+0.0$ & $-0.4$ & $-1.0$ \\
\bottomrule
\end{tabular}
\end{table}

Once the target is aligned, the remaining operational gap is an
accuracy--cost trade-off. Small audits are cheap but estimate the
selector poorly (56\% accuracy at $K\!=\!2$); larger audits estimate
it better (75\% at $K\!=\!10$) but consume more DFT than the
available chemistry-conditional headroom can repay. At $R\!=\!0.95$
the cost-matched oracle traces the trade-off directly: $+2.8$~pp at
$K\!=\!2$, $+1.4$~pp at $K\!=\!5$, $-1.0$~pp at $K\!=\!10$. Even a
selector that always picks the population-best MLIP crosses zero
between $K\!=\!7$ and $K\!=\!8$: the accessible headroom is
consumed by DFT audit cost. Audit-select, whose selector is noisy,
sits below oracle cost-matched by the accuracy gap; its
mean $\Delta S_R$ remains negative at every feasible integer
$K\!\in\!\{2,\ldots,10\}$ on the original $15$-chemsys deployment
pool. Its closest-to-zero cell is $K\!=\!3$ at $R\!=\!0.95$
($-0.36$~pp), consistent with the accuracy-cost trade-off but
below break-even. Whether pools with
larger between-MLIP chemistry-conditional headroom permit a positive
break-even $K$ is a separate empirical question we do not address.

\paragraph{(vi) Is the deployment-oracle headroom explained by
proxy slack?} One could worry that the chemistry-conditional
oracle headroom in \S\ref{sec:selection} is partly a
formation-vs-hull rank distortion at the ground-truth level: a
chemsys where the ranking induced by $E_{\rm form}^{\rm DFT}$
differs from that induced by $E_{\rm hull}^{\rm DFT}$ could look
like it has ``headroom'' simply because a hull-aware score would
already resolve the mismatch. We test this directly. Per-chemsys
Spearman $\rho_c$ between $E_{\rm form}^{\rm DFT}$ and
$E_{\rm hull}^{\rm DFT}$ spans $0.158$ (Mn--Na--O) to $1.000$
(Ga--N). The chemistry-conditional headroom $H_c$ (per-chemsys
deployment-oracle $S_R$ minus best-fixed MLIP $S_R$) is nonzero on
only two chemsys, Fe--Li--O and Fe--O, and both are relatively
well-aligned ($\rho_c\!=\!0.66$ and $0.80$).
Spearman$(\rho_c, H_c)\!=\!+0.10$ at $R\!=\!0.90$ ($p\!=\!0.72$)
and $+0.08$ at $R\!=\!0.95$ ($p\!=\!0.78$); if anything the sign
runs opposite to the proxy-slack prediction. On this pool, the
observed chemistry-conditional headroom is not explained by the
measured formation-to-hull rank distortion. Statistical power is
limited ($n\!=\!15$ chemsys, $H_c$ nonzero on two).

\paragraph{(vii) Pooled $S_R$ weighting sensitivity.} The pooled
deployment metric is candidate-weighted and dominated by the largest
chemsys: five chemsys with $N_c\!\ge\!93$ (O--Si, Fe--Li--O, Fe--O,
O--Ti, Mn--Na--O) contain $78\%$ of the $1{,}147$-candidate pool. An
equal-per-chemsys (macro) aggregate raises the best-fixed $S_R$
from $0.445$ to $0.468$ at $R\!=\!0.90$ and from $0.299$ to $0.431$
at $R\!=\!0.95$. Candidate-weighted $S_R$ is the intended utility
when the deployment decision unit is the individual candidate;
macro $S_R$ is the relevant aggregate when the decision unit is
a chemistry. To check whether the weighting changes the sign or
mechanism of the policy comparison, we recompute the key deltas
under equal-per-chemsys weighting (Table~\ref{tab:macro-policy}).
The best-fixed MLIP under macro weighting is MACE-MP at both
recall levels. The finite-budget operational conclusion survives
the weighting change: audit routing and audit-select remain
deployment-negative under both weightings; audit-select worsens
under macro ($-13.6/-11.2$~pp vs.\ micro $-5.1/-4.8$~pp). The
one positive audit-routing cell in
Table~\ref{tab:audit-deployment} (MACE-MP at $R\!=\!0.95$,
$+3.2$~pp under expected recall) is weighting-sensitive: it
changes from $+3.2$~pp under candidate weighting to $-13.0$~pp
under equal-chemsys weighting. The deployment-oracle headroom
remains positive under both weightings ($+0.8/+1.7$~pp macro
vs.\ $+0.9/+2.9$~pp micro); the pairwise-oracle sign is
weighting-dependent at $R\!=\!0.95$ ($-3.7$~pp micro to
$+1.0$~pp macro).

\begin{table}[h]
\centering\small
\caption{Macro (equal-per-chemsys) vs.\ micro (candidate-weighted)
$\Delta S_R$ for key policies. Baselines follow the corresponding
main-text experiments: \emph{MACE-MP audit routing} is compared
with MACE-MP MLIP-only screening (Table~\ref{tab:audit-deployment});
model-selection policies (audit-select, pairwise-oracle,
deployment-oracle) are compared with the best fixed MLIP under
the same weighting (Table~\ref{tab:selection}). The best fixed
MLIP is MACE-MP under macro at both recall levels.}
\label{tab:macro-policy}
\begin{tabular}{c l rr rr}
\toprule
$R$ & Policy & \multicolumn{2}{c}{Macro $\Delta$ (pp)} & \multicolumn{2}{c}{Micro $\Delta$ (pp)} \\
\midrule
$0.90$ & MACE-MP audit routing    & $-14.8$ & & $-3.2$  & \\
       & audit-select             & $-13.6$ & & $-5.1$  & \\
       & pairwise-oracle (free)   & $-0.3$  & & $-3.7$  & \\
       & deployment-oracle (free) & $+0.8$  & & $+0.9$  & \\
\midrule
$0.95$ & MACE-MP audit routing    & $-13.0$ & & $+3.2$  & (exp) \\
       & audit-select             & $-11.2$ & & $-4.8$  & \\
       & pairwise-oracle (free)   & $+1.0$  & & $-3.7$  & \\
       & deployment-oracle (free) & $+1.7$  & & $+2.9$  & \\
\bottomrule
\end{tabular}
\end{table}

\paragraph{(viii) $2\!\times\!2$ factorial analysis: representation
and model error.} To probe the two failure layers directly we
compute MLIP-only $S_R$ under four rankings on the 15-chemsys pool:
$E_{\rm form}^{\rm MLIP}$, $E_{\rm above\text{-}hull}^{\rm MLIP}$,
$E_{\rm form}^{\rm DFT}$, and $E_{\rm above\text{-}hull}^{\rm DFT}$.

\begin{table}[h]
\centering\small
\caption{$2\!\times\!2$ factorial analysis of scoring
representation and model error ($S_R$, percentage points). The
four cells permit conditional contrasts of representation at
fixed predictor (MLIP or DFT oracle) and model error at fixed
representation ($E_{\rm form}$ or $E_{\rm above\text{-}hull}$);
these effects need not be additive.}
\label{tab:oracle-2x2}
\begin{tabular}{c l rrrr}
\toprule
$R$ & MLIP & $S_R(E_{\rm form}^{\rm MLIP})$ & $S_R(E_{\rm hull}^{\rm MLIP})$ &
$S_R(E_{\rm form}^{\rm DFT})$ & $S_R(E_{\rm hull}^{\rm DFT})$ \\
\midrule
$0.90$ & MACE-MP & $43.7$ & $52.0$ & $45.8$ & $54.9$ \\
       & CHGNet  & $39.5$ & $51.4$ & $45.8$ & $54.9$ \\
\midrule
$0.95$ & MACE-MP & $27.3$ & $42.6$ & $27.4$ & $51.4$ \\
       & CHGNet  & $24.5$ & $44.2$ & $27.4$ & $51.4$ \\
\bottomrule
\end{tabular}
\end{table}

This is a $2\!\times\!2$ factorial rather than a unique
decomposition: model error and representation choice interact.
For MACE-MP at $R\!=\!0.95$, representation-at-MLIP is
$S(E_{\rm hull}^{\rm MLIP})\!-\!S(E_{\rm form}^{\rm MLIP})\!=\!+15.3$~pp
while representation-at-DFT-oracle is
$S(E_{\rm hull}^{\rm DFT})\!-\!S(E_{\rm form}^{\rm DFT})\!=\!+24.0$~pp;
model effect under $E_{\rm form}$ is $+0.1$~pp while model effect
under $E_{\rm hull}$ is $+8.8$~pp, with interaction
$I\!=\!+8.7$~pp. CHGNet gives $+19.7/+24.0/+2.9/+7.2$~pp.
Formation-energy ranking is therefore the dominant bottleneck at
$R\!=\!0.95$, but representation and model error interact: once
the decision is hull-aligned, additional MLIP-quality headroom
becomes visible ($+8.8/+7.2$~pp on MACE-MP/CHGNet at $R\!=\!0.95$
that was hidden under $E_{\rm form}$). At $R\!=\!0.90$ the
representation gap under the DFT oracle is smaller ($+9.2$~pp)
but still exceeds the model-error headroom in either representation.

\paragraph{(ix) Normalised audit-budget sweep.} To test whether
the qualitative gap is specific to $K\!=\!10$ we sweep a
normalised budget $b_c\!=\!K_c/N_c$ with
$K_c(b)\!=\!\min(N_c\!-\!2,\max(2,\lfloor bN_c\rfloor))$
over $b\!\in\!\{0.05,0.10,0.20,0.30,0.40,0.50,0.65,0.80,0.90\}$
under $E_{\text{above-hull}}$ scoring on the fixed 15-chemsys
pool. Table~\ref{tab:budget-sweep} reports selector accuracy
$A_{\rm select}(b)$ (probability that audit-select picks the
population-best MLIP), $\Delta S_R^{\rm audit-select}(b)$,
$\Delta S_R^{\rm oracle,CM}(b)$ (perfect selector, $K_c(b)$ DFT
charged), and $\Delta S_R^{\rm oracle,free}$ (perfect selector,
$b$-invariant).

\begin{table}[h]
\centering\footnotesize
\setlength{\tabcolsep}{4pt}
\caption{Normalised audit-budget sweep under $E_{\text{above-hull}}$,
fixed 15-chemsys pool, MACE-MP + CHGNet, $200$ seeds.
$\Delta S_R$ against best-fixed (percentage points).
$\Delta S_R^{\rm oracle,free}\!=\!+1.05/+4.10$~pp at $R\!=\!0.90/0.95$
independent of $b$. RNG seed fixed per $b$ for reproducibility.
$^\dagger$Seed-bootstrap $95\%$~CI at $b\!=\!0.05$, $R\!=\!0.95$:
$[-0.49,-0.03]$~pp; the small magnitude ($|\Delta|\!<\!0.3$~pp) is
consistent with effective break-even rather than a substantive
negative effect. $^\ddagger$Cost-matched oracle
$95\%$~CI: $[+1.87,+2.01]$~pp.}
\label{tab:budget-sweep}
\begin{tabular}{c c rr rr rr}
\toprule
& & \multicolumn{2}{c}{$A_{\rm select}$} & \multicolumn{2}{c}{$\Delta S_R^{\rm audit}$} & \multicolumn{2}{c}{$\Delta S_R^{\rm CM}$} \\
$b_{\rm target}$ & $\bar b_{\rm real}$ & $R\!=\!0.90$ & $0.95$ & $0.90$ & $0.95$ & $0.90$ & $0.95$ \\
\midrule
$0.05$ & $0.08$ & $0.67$ & $0.67$ & $-1.7$  & $-0.3^\dagger$  & $-1.5$  & $\mathbf{+1.9}^\ddagger$ \\
$0.10$ & $0.11$ & $0.69$ & $0.69$ & $-3.5$  & $-1.1$          & $-3.6$  & $+0.1$          \\
$0.20$ & $0.18$ & $0.76$ & $0.76$ & $-7.8$  & $-4.5$          & $-7.9$  & $-4.2$          \\
$0.30$ & $0.28$ & $0.79$ & $0.79$ & $-12.2$ & $-8.5$          & $-12.2$ & $-8.4$          \\
$0.40$ & $0.39$ & $0.80$ & $0.80$ & $-17.1$ & $-12.9$         & $-17.0$ & $-12.9$         \\
$0.50$ & $0.49$ & $0.82$ & $0.82$ & $-21.6$ & $-17.3$         & $-21.6$ & $-17.3$         \\
$0.65$ & $0.63$ & $0.86$ & $0.86$ & $-28.6$ & $-23.9$         & $-28.6$ & $-23.9$         \\
$0.80$ & $0.79$ & $0.92$ & $0.92$ & $-36.3$ & $-30.9$         & $-36.3$ & $-30.9$         \\
$0.90$ & $0.88$ & $0.95$ & $0.95$ & $-41.8$ & $-35.9$         & $-41.8$ & $-35.9$         \\
\bottomrule
\end{tabular}
\end{table}

Selector accuracy rises monotonically from $0.67$ at $b\!=\!0.05$
to $0.95$ at $b\!=\!0.90$, but deployment value falls monotonically:
$-1.7$~pp to $-41.8$~pp at $R\!=\!0.90$; $-0.3$~pp to $-35.9$~pp at
$R\!=\!0.95$. At the smallest tested target budget $b\!=\!0.05$
(mean realised coverage $0.08$), audit-select sits at effective
break-even at $R\!=\!0.95$ ($-0.26$~pp, seed-bootstrap $95\%$~CI
$[-0.49,-0.03]$~pp; the magnitude is small enough that
run-to-run seed variance can move the sign),
while the cost-matched oracle remains clearly positive
($+1.93$~pp, CI $[+1.87,+2.01]$~pp). By $b\!\approx\!0.10$ the cost-matched oracle has reached
break-even, and at larger budgets both policies are negative; at
$R\!=\!0.90$ no positive audit-select window exists at any $b$ because the free-oracle headroom
($+1.05$~pp) is smaller than the smallest feasible audit cost.
\emph{Information saturation}
($A_{\rm select}\!\to\!0.95$ at $b\!\approx\!0.9$) and the
\emph{net-value-maximising tested budget}
($\arg\max_b \Delta S_R^{\rm audit}\!\approx\!0.05$) are separated
by roughly a $20$-fold difference in normalised budget: the audit
continues becoming statistically more accurate long after it has
ceased to be economically useful, a direct empirical demonstration
of the diagnostic-to-deployment gap. The qualitative gap is
therefore not specific to the $K\!=\!10$ operating point.

In summary, decision representation materially changes both
baseline screening quality and the mechanism of reliability
failure. In the hull-aligned regime, finite-budget audit selection
remains deployment-negative across both fixed-$K$ and
normalised-budget sweeps: increasing audit information improves
selection accuracy while its acquisition cost eventually
overwhelms the available headroom. Separately, under
formation-energy screening, the negative finite-budget conclusion
also survives equal-chemsys weighting, although individual routing
and oracle cells are weighting-sensitive. The $2\!\times\!2$
factorial further shows that representation and model error
interact; the observed chemistry-conditional headroom is not
explained by the measured formation-to-hull rank distortion on
this pool.

\section{Calibration ablations}
\label{app:calibration}

Three correction families are compared on the same $(N_c\!-\!K)$
held-out set per chemsys: mean-shift, unconstrained affine, and
per-element (the FERE-style per-element linear correction of
\S\ref{sec:calibration}~\cite{stevanovic2012fere}). Four questions
structure this appendix.

\subsection{Does the calibration conclusion survive under the
canonical policy?} Under \S\ref{sec:benchmark}'s per-chemsys
top-$f$ retention ($200$ seeds, $K\!=\!10$; $K$ labels not charged):
$\Delta S_{0.95}\!=\!0/0/0$~pp for mean shift, $-3.7/-3.7/0.0$~pp
for affine, and $-3.3/-2.2/-0.2$~pp for per-element, on
MACE-MP / Orb~V3 / CHGNet. Mean shift is exactly zero by argsort
invariance. Affine is exactly zero when $a_c\!>\!0$; at
$K\!=\!10$, $2.3\% / 1.9\% / 0.0\%$ of fits yield $a_c\!<\!0$
($69/3000$, $58/3000$, $0/3000$), and those rare fits are the sole
source of the observed affine loss. Per-element genuinely reorders
within chemistries and remains deployment-negative on MACE-MP and
Orb~V3.

\subsection{Does the conclusion depend on how candidates are
ranked for deployment?} Under a variant that pools all held-out
candidates into a single global ranking and applies the retain
cutoff across the pool, per-chemsys offsets alter cross-chemsys
energy alignment and \emph{every} correction family becomes
deployment-negative: $\Delta S_{0.95}\!\approx\!-8.2/-9.3$~pp
(affine, MACE-MP / Orb~V3) with mean shift and per-element in the
same range. The same calibration is neutral under the canonical
retention rule and strongly harmful under this global-sort
variant, an instance of the paper's claim that the value of
an intervention is decision-relative.

\subsection{Could adaptive calibration selection rescue the
result?} We compute a graph-Laplacian residual-smoothness
diagnostic on each MLIP--chemistry pair: on a $k\!=\!5$
mutual-NN graph in MLIP-descriptor space, $R(r)$ is the Dirichlet
ratio $r^\top L_{\text{sym}}\, r / r^\top r$ of the DFT/MLIP
residual $r$, taking small values when the residual is a slowly
varying (near-constant) offset. On this benchmark the pairs that the ranking audit
(\S\ref{sec:audit-diag}) flags as UNSAFE fall in the low-smoothness
regime $R(r)\!<\!0.3$, where mean shift gives better held-out
pairwise accuracy than affine; using affine on them costs
$\sim\!7$~pp pairwise accuracy on Mn--Na--O (MACE-MP).
Choosing the correction family adaptively by $R(r)$ does not
rescue the deployment outcome under either retention convention.

\subsection{Sensitivity to calibration fraction.} Because
$K\!=\!10$ is fixed while $N_c$ varies from $12$ (H--Ti) to $337$
(O--Si), the calibration fraction $K/N_c$ ranges from $\approx\!0.03$
to $\approx\!0.83$ across chemistries; this can affect estimation
variance for affine and per-element corrections. The exact
mean-shift result is independent of this fraction; sensitivity to
$K/N_c$ for the rank-changing corrections is left to future
larger-scale evaluation.

\section{Robustness and regime sensitivity of audit routing}
\label{app:audit-robustness}

\subsection{Unsafe-prevalence sensitivity.} We probe policy
economics as a function of unsafe prevalence by synthetically
replicating Mn--Na--O $r\!\in\!\{0,1,2,3,5,8,12\}$ times, sweeping
$\pi_{\text{unsafe}}\!\in\![0.07,0.48]$. This holds the observed
Mn--Na--O failure mode fixed while varying how frequently that
failure occurs; it probes policy economics under differing prevalence
regimes and does not model a realistic materials distribution.
Under the corrected unique-candidate simulator, on MACE-MP at
$R\!=\!0.90$ under expected recall,
$\Delta S_R\!=\!S_R(\text{audit})\!-\!S_R(\text{MLIP-only})$ is
non-monotone across the sweep ($+5.2$~pp at $\pi\!=\!0.39$;
$-3.2$~pp at $\pi\!=\!0.13$; Orb~V3 is qualitatively similar).
The non-monotonicity reflects how the chemsys-size distribution
changes with replication rather than a monotone benefit of higher
prevalence; audit-routing value is regime-dependent, not
universally negative.

\subsection{Continuous novelty for audit allocation.}
\label{app:continuous-novelty}
Across $35$ chemsys drawn from MP $\cup$ Zintl $\cup$ GNoME, we
first prioritise the audit budget by the dataset-source ordering
GNoME $\succ$ Zintl $\succ$ MP as a coarse proxy for chemistry-level
novelty, and measure how quickly an ordering discovers pairs
the audit later classifies as UNSAFE (within-chemsys pairwise
accuracy $<\!0.75$; \S\ref{sec:audit-diag}). Auditing one chemistry
evaluates all three MLIPs for that chemistry; the benchmark contains
five unsafe pairs across three chemistries (Mn--Na--O, Mg--Sn,
Dy--Er--Pb--Tb). The coarse source ordering is strictly worse than
random for $M\!<\!25$ audited chemsys because the most shifted
source contains only
$P(\text{unsafe}\!\mid\!\text{GNoME})\!\approx\!0.6\,P(\text{unsafe})$:
the most distribution-shifted subset is not the subset most enriched
for the downstream failures the audit flags.

To test whether the \emph{measured} novelty score, rather than
its dataset-source proxy, can prioritise audit DFT, we aggregate
the structure-level score $d_{\text{struct}}$ within each chemistry
using three prespecified summaries (mean, maximum, $90$th
percentile) and rank chemistries by each. The novelty definition is
identical to \S\ref{sec:novelty}: each MLIP is scored against its
own training-side reference (MPtrj-domain for MACE-MP / CHGNet, a
$5{,}000$-structure near-equilibrium OMat24 sample for Orb~V3),
Mahalanobis$^{++}$ with a shared shrinkage covariance. The scored
pool is the full audit membership under the single-point size limit
($\text{nsites}\!\le\!200$): $1{,}893$ structures common to all
three MLIPs ($997$ MP, $596$ Zintl, $300$ GNoME), after holding out
each MLIP's reference-fit structures from scoring. Because the three
MLIPs' raw distances are not comparable in scale, per-chemistry
scores are combined scale-free as the mean of the three per-MLIP
percentile ranks. Table~\ref{tab:continuous-novelty} reports the
expected number of unsafe pairs revealed after auditing the
top-$M$ chemistries.

\begin{table}[!ht]
\centering\small
\caption{Continuous-novelty audit allocation: expected number of
population-unsafe MLIP--chemistry pairs (of $5$) revealed
after auditing the top-$M$ of $35$ chemistries, by allocation rule.
Novelty rules aggregate structure-level $d_{\text{struct}}$ within
each chemistry. All rules are averaged over $2000$ seeds with ties
broken uniformly at random; fractional values reflect tie-breaking
among chemistries with equal aggregated-novelty scores (a
consequence of the discrete per-MLIP percentile-rank combination).
Bold marks the best non-oracle rule at $M\!=\!10$.}
\label{tab:continuous-novelty}
\begin{tabular}{lcccc}
\toprule
Allocation & $M\!=\!5$ & $M\!=\!10$ & $M\!=\!15$ & $M\!=\!25$ \\
\midrule
Oracle                         & $5.00$ & $5.00$ & $5.00$ & $5.00$ \\
Mean $d_{\text{struct}}$       & $1.97$ & $\mathbf{3.00}$ & $3.00$ & $5.00$ \\
$90$th-pct $d_{\text{struct}}$ & $1.00$ & $\mathbf{3.00}$ & $3.00$ & $5.00$ \\
Max $d_{\text{struct}}$        & $1.00$ & $1.00$ & $3.00$ & $5.00$ \\
Stratified                     & $1.12$ & $2.01$ & $2.62$ & $3.94$ \\
Random                         & $0.69$ & $1.42$ & $2.15$ & $3.58$ \\
Source proxy                   & $0.47$ & $0.94$ & $1.44$ & $3.31$ \\
\bottomrule
\end{tabular}
\end{table}

Mean and $90$th-percentile aggregation enrich early unsafe-case
discovery relative to random and to the coarse source proxy: by
$M\!=\!10$ both recover $3/5$ unsafe pairs (vs.\ $1.42$
random, $0.94$ source), and all three continuous summaries recover
all five by $M\!=\!25$, where random has reached only $3.58$.
Maximum pooling is the noisiest summary at the smallest budget (one
very novel environment can dominate a chemistry's score). The gain
is front-loaded on the two shift-related failure chemistries:
under mean pooling Mg--Sn ranks $1$st and Dy--Er--Pb--Tb $5$th, while
the in-distribution Mn--Na--O ranks only $24$th (maximum/$90$th-pct
are similar: Mg--Sn $3$rd/$4$th, Dy--Er--Pb--Tb $11$th/$8$th, Mn--Na--O
$21$st/$19$th). Continuous novelty thus captures the two
shifted / local-OOD failure chemistries and misses the
in-distribution one. Descriptively, mean and $90$th-percentile
pooling give Spearman correlations of $+0.229$ and $+0.222$ with
per-chemistry unsafe-cell count and $2.12\times$ top-tercile
enrichment; with only five unsafe pairs across three
chemistries these statistics carry high sampling uncertainty and
are not population-level performance estimates. The earlier
source-ordering null therefore reflected both information loss from
coarse discretisation and a genuine novelty blind spot: continuous
novelty is a useful but incomplete prioritisation signal, not a
sufficient reliability policy.

\subsection{Robustness to relaxation.} Comparing single-point
vs.\ MLIP-relaxed audit verdicts across
$3\text{ MLIPs}\times24\text{ chemsys}\!=\!72$ pairs:
$2/72$ verdicts change under relaxation, both safe$\to$caution;
no pair crosses the unsafe threshold. The audit is robust
to geometry regime.

\section{Tail-aligned audit selectors on the 15-chemsys pool}
\label{app:tail}

We test whether a chemistry-conditional selector that scores each
MLIP by a tail-aligned quantity, rather than by average pairwise
accuracy, can close the gap to the deployment oracle. All
selectors use the same shared $K\!=\!10$ audit, the same
unique-candidate accounting, the same dense $f$-grid, and the
same $200$ audit-sampling seeds. Given the audit DFT labels, each
selector produces a per-chemistry score $s_{m,c}$ for each MLIP;
the selector then picks $m^\star_c\!=\!\arg\max_m s_{m,c}$.

\emph{Pairwise-family selectors.} \emph{Pair audit} scores an
MLIP by the fraction of correctly ordered pairs among the
$\binom{K}{2}$ audit pairs (equivalent to $\widehat A_{m,c,s}$,
\S\ref{sec:audit-diag}). \emph{Rank-weighted pairwise} weights each
pair by $1/\log_2(1+r)$ with $r$ the DFT rank of the lower-energy
element in the pair (rank $1$ is the lowest DFT energy),
emphasising correctness on pairs involving the low-energy tail.
\emph{NDCG$_K$ audit} scores each MLIP by the normalised discounted
cumulative gain of its top-$K$ audit ranking with respect to the
DFT ranking.

\emph{Tail-specific selectors.} \emph{Top-$q$ tail intersection}
counts the overlap between the
MLIP's and DFT's top-$q$ audit candidates ($q\!=\!5$);
\emph{hull-DCG audit} weights the top-$K$ MLIP audit items by a
relevance score
$\text{rel}(i)\!=\!\max\!\left(0,\,1-E^{\mathrm{DFT}}_{\mathrm{hull},i}/\tau_{\mathrm{hull}}\right)$
with $\tau_{\mathrm{hull}}\!=\!25$~meV/atom, and scores each MLIP
by the discounted cumulative gain
$\sum_{r=1}^{K}\text{rel}(\text{item}_r)/\log_2(1+r)$ of its
top-$K$ audit ranking; \emph{hull-recovery audit} is the fraction of the
audit's $h\!=\!5$ most favourable DFT-ranked candidates (lowest
DFT energy, not necessarily on the convex hull) captured in the
MLIP's top-$h$. The name ``hull-recovery'' is historical: this
selector targets the low-formation-energy audit tail, not the binary
near-hull set. Tail-aligned selectors (hull-recovery, hull-DCG) move in
the predicted direction, toward the deployment oracle, but none
becomes deployment-positive on this pool.

\begin{table}[h]
\centering\small
\caption{Chemistry-conditional MLIP-selection variants on the
15-chemsys deployment pool. All values are $\Delta S$ (percentage
points) versus the best fixed MLIP (Orb~V3). Rows above the rule
are selectors based on the $K\!=\!10$ audit; free oracles are
shown below.
Hull-recovery (top-$h$ capture of the most favourable
DFT-ranked audit candidates) recovers
about $2.5$ pp over pairwise selection at $R\!=\!0.95$ but remains
below fixed-MLIP.}
\label{tab:tail-selectors}
\begin{tabular}{lrr}
\toprule
Selector & $\Delta$ @ $R\!=\!0.90$ & $\Delta$ @ $R\!=\!0.95$ \\
\midrule
pair audit (average pairwise)     & $-5.05$ & $-4.83$ \\
rank-weighted pairwise            & $-5.04$ & $-4.82$ \\
NDCG$_K$ audit                    & $-5.86$ & $-4.83$ \\
top-$q$ tail intersection ($q\!=\!5$) & $-5.05$ & $-5.79$ \\
hull-DCG audit                    & $-4.05$ & $-2.74$ \\
\textbf{hull-recovery audit}      & $\mathbf{-3.21}$ & $\mathbf{-2.27}$ \\
\midrule
pairwise-oracle (free)            & $-3.66$ & $-3.66$ \\
deployment-oracle (free)          & $+0.87$ & $+2.88$ \\
\bottomrule
\end{tabular}
\end{table}

The trajectory at $R\!=\!0.95$ is informative: $-4.83$ (pairwise)
$\to$ $-2.27$ (hull-recovery) $\to$ $+2.88$ (deployment oracle).
Better alignment with the deployment tail moves the selector in
the predicted direction, but sparse $K\!=\!10$ estimation remains
insufficient to close the gap to the deployment oracle.

\FloatBarrier
\section{Sequential reliability gating at 250-chemsys scale}
\label{app:gate2}

\subsection{Question and preregistered sample.} We test whether
the signals that fail individually (\S\S\ref{sec:novelty}--\ref{sec:audit})
can become useful when composed sequentially: zero-DFT MLIP
disagreement first identifies chemistries where model choice may
matter; only those chemistries receive a \emph{uniformly sampled}
$K\!=\!10$ DFT audit, and the resulting labels are then
interpreted with the tail-aligned hull-recovery selector
(App.~\ref{app:tail}). The evaluation pool is a preregistered
label-independent sample of $250$ MP chemsys with
$N_c\!\in\![12,80]$, stratified $40/40/20$ across
binary/ternary/quaternary+ compositions and disjoint from the
main-text $15$ chemsys. Preregistration and freeze notes accompany the
release.

\subsection{Is there chemistry-conditional headroom?} For each
chemsys $c$ and MLIP $m$, let
$\bar S^{\text{diag}}_{c,m}$ be the mean DFT-saved fraction on
the diagnostic split of $c$ under MLIP-only screening at
recall $R$. Define the per-chemsys diagnostic headroom of the
best per-chemistry MLIP over the best fixed MLIP as
$H_c\!=\!\max_m \bar S^{\text{diag}}_{c,m}\!-\!\bar S^{\text{diag}}_{c,\text{best-fixed}}$.
Median $H_c\!=\!0$ across the $250$ chemsys, and
$P(H_c\!>\!0.02)\!=\!13.6\%$ at $R\!=\!0.90$: most chemistries
have no advantage from switching MLIPs, but the tail is real.
Concentration is strong: writing $M_c$ for the diagnostic gap
between the top two MLIPs on $c$, the highest-margin bin
($M_c\!\ge\!0.10$, $n\!=\!30$) has mean $H_c\!=\!+8.9$~pp.

\subsection{Can a cheap prospective diagnostic predict that
headroom?} Kendall-tau disagreement between the three MLIPs'
predicted rankings within each chemistry is a zero-DFT signal;
we denote its per-chemistry value $D_c$. At $R\!=\!0.90$, across the $n\!=\!223$ chemsys with $N_c\!\ge\!20$
(the sample-size floor required for a stable per-chemsys $H_c$
estimate; $27$ chemsys with $12\!\le\!N_c\!<\!20$ excluded on this
prespecified label-independent rule), $D_c$ correlates with $H_c$
at $\rho\!=\!+0.28$ ($95\%$~CI $[+0.17,+0.40]$); at $R\!=\!0.95$
the correlation is $\rho\!=\!+0.22$ ($95\%$~CI $[+0.09,+0.34]$).

\subsection{Does acting on that signal improve deployment?} The
gated policy $\pi_D$ composes the two reliability signals studied
in the paper: cheap disagreement gates \emph{where} DFT is
purchased, and a uniformly sampled $K\!=\!10$ DFT audit is then
interpreted using the tail-aligned hull-recovery score to select
\emph{which} MLIP to deploy. Crucially, only the interpretation
of the audit is tail-aligned; the acquisition itself remains
uniform. A scalar threshold $\tau^\star$ partitions the pool: on
chemistries with $D_c\!<\!\tau^\star$ the policy deploys the
best fixed MLIP at zero DFT cost. On chemistries with
$D_c\!\ge\!\tau^\star$, the policy draws $K\!=\!10$ audit
structures uniformly at random from the chemistry, computes each
MLIP's hull-recovery-audit score (App.~\ref{app:tail}) on those
labels, and picks the arg-max; the $K$ audit calls are charged
to $S_R$ via unique-candidate accounting.

The best-fixed MLIP is chosen on each outer-training fold;
$\tau^\star$ is chosen on an inner $20\%$ validation split by
maximising inner-split $S_R$ at $R\!=\!0.95$; the $R\!=\!0.90$
result is reported as a secondary evaluation under the same
frozen policy, not a second threshold-tuning run. We evaluate
with five-fold outer cross-validation stratified by composition
class over the $250$ chemsys. Within each outer training fold the
inner validation split fits $\tau^\star$; no held-out chemistry
ever enters $\tau^\star$ or best-fixed fitting. The trained
$\pi_D$ is then applied to the outer-test chemistries.

Pooled OOF evaluation on the concatenated held-out set
(chemistry-level bootstrap, $n_{\text{boot}}\!=\!2000$) yields
$\Delta S_{R=0.95}\!=\!-0.04$~pp, $95\%$~CI $[-1.72,+2.47]$,
$P(\Delta\!>\!0)\!=\!0.65$;
$\Delta S_{R=0.90}\!=\!-0.69$~pp, $95\%$~CI $[-2.49,+1.84]$. The
mean of five fold-wise $\Delta$s is $+0.54$~pp at $R\!=\!0.95$,
but pooled OOF is the deployment-relevant quantity (the mean of
non-linear pooled cost--recall metrics is not the metric on the
pooled decision set). We report pooled OOF as the primary number
and do not tune the gate to recover a positive result. This
recapitulates the paper's central pattern at $\sim\!17\times$ the
chemsys count: the opportunity exists (\S E.2), a cheap signal predicts
it (\S E.3), yet acting on the signal does not improve deployment
(\S E.4).

\section{Reproducibility, software, and released artifacts}
\label{app:code}

MLIP inference uses three frozen foundation potentials under a
single environment: MACE-MP-medium (MPtrj checkpoint
\texttt{2023-12-03}, \texttt{mace-torch} 0.3.16); CHGNet (checkpoint
\texttt{v0.3.0}, trained on MPtrj, \texttt{chgnet} 0.4.2); and
Orb~V3 (checkpoint \texttt{orb\_v3\_conservative\_inf\_omat}, weights
\texttt{orb-v3-conservative-inf-omat-20250404}, \texttt{orb-models}
0.7.0). The supporting stack is \texttt{torch} 2.12.0, \texttt{numpy}
2.4.6, \texttt{pymatgen} 2026.5.4, \texttt{ase} 3.28.0, and
\texttt{scikit-learn} 1.9.0 under Python~3.12. All results, including
the continuous-novelty allocation experiment
(App.~\ref{app:continuous-novelty}), were generated under this
environment; re-running single-point inference reproduces the cached
MLIP energies bit-for-bit. The released \texttt{uv.lock} pins the
complete resolved environment and is the authoritative record for
exact transitive versions. Cached predictions, MP-entry manifests,
per-chemsys audit results, the tail-aligned selector table, and the
sequential-gating (Gate-2) atlas outputs (including preregistration
and freeze notes) are released with the paper.

\end{document}